# Semantic Similarity Strategies for Job Title Classification

Yun Zhu, Faizan Javed, Ozgur Ozturk
Careerbuilder LLC.
5550-A Peachtree Pkwy
Norcross, GA 30092
{yun.zhu, faizan.javed, ozgur.ozturk}@careerbuilder.com

## ABSTRACT

Automatic and accurate classification of items enables numerous downstream applications in many domains. These applications can range from faceted browsing of items to product recommendations and big data analytics. In the online recruitment domain, we refer to classifying job ads to pre-defined or custom occupation categories as job title classification. A large-scale job title classification system can power various downstream applications such as semantic search, job recommendations and labor market analytics. In this paper, we discuss experiments conducted to improve our in-house job title classification system. The classification component of the system is composed of a two-stage coarse and fine level classifier cascade that classifies input text such as job title and/or job ads to one of the thousands of job titles in our taxonomy. To improve classification accuracy and effectiveness, we experiment with various semantic representation strategies such as average W2V vectors and document similarity measures such as Word Movers Distance (WMD). Our initial results show an overall improvement in accuracy of Carotene[1].

## 1. INTRODUCTION

Many e-commerce and web properties have a need to automatically classify millions of items to thousands of categories with a high level of accuracy. Such large-scale item classification systems have many downstream applications such as product recommendations, faceted search, semantic search and big data analytics. In the online recruitment domain, classification of job ads can power applications such as labor market analytics, job recommendations, and semantic search. We refer to classifying job ads (text documents composed of title, description and requirements fields) to pre-defined or custom occupation categories as job title classification. For automatic job title classification, we developed a system called Carotene, which has: i) a taxonomy discovery component that leverages clustering techniques to discover job titles from data sets to create a custom job title taxonomy, and ii) a two stage coarse and fine level classifier that uses an SVM-KNN cascade to classify input text to the most appropriate job titles in our custom taxonomy. More details on the system can be found in [1]. The focus of this paper is on improvements based on semantically rich document representations (e.g., using enriching vectors) and similarity measures (e.g., WMD) for the classification component of the system. These semantic enrichment strategies replace the bag of words (BOW) representation that is most commonly used in text classification with semantically related terms (or concepts) derived from resources that establish relations between entities and concepts. Semantic representations are more adept than BOW representations at handling synonyms, polysemous words and multi-word expressions. In this paper, we show our experiment results with job title representation using Word2Vec, Doc2Vec and WMD. The rest of the paper is arranged as follows. Section 2 discusses related work on semantic enrichment strategies for text classification as well as their application in various domains. Section 3 briefly describe the three methods we experiment with. Section 4 explains the use case and shows the performance results as well.

## 2. RELATED WORK

Enriching vectors and semantic kernels are the two most commonly used semantic enrichment techniques for text classification. In the enriching vectors approach [2], a document representation is enriched with some or all of the following: hypernyms, synonyms and related concepts. Semantic kernels [3] leverage a semantic proximity matrix to transform document representations into linearly separable semantic representations. Semantic kernels are usually used with SVMs and applied before classifier training time. In [4], enriching vectors and semantic kernels were used to classify medical text. Their observation was that enriching vectors performed better than BOW-based approaches while semantic kernels degraded performance by introducing noise in document representations. However, [5] details a semantic kernel approach that actually improved the performance of the SVM algorithm when the dimensionality of the input feature space is large and training data is scarce. For computing similarities between documents, the WMD distance [6] has recently been shown to provide lower classification errors when used with distance-based classifiers. For job title classification, LinkedIn uses a phrase-based classification system that relies on the near-sufficiency property of short text [7]. The property of near-sufficiency implies that short text documents usually contain more information about the class label of the document than long text. Consequently, this approach considers job titles only for classification and does not make use of other job ad fields such as description and requirement. A semantic enrichment approach to job title classification is discussed in [8]. This approach semanti-cally enriches job categories with contextually relevant terms derived from a corpus of job ads. A field-to-field similarity matching approach then matches job ads to job categories.

## 3. METHODS

### 3.1 Baseline

Previous version of Carotene, our machine learning-based, semi-supervised, multi-class job title classification system, consisted of a two-stage cascade of coarse and fine grade classifiers. Coarse-grade classifier assigns the title to one of the 23 top level categories, called SOC majors, after the Standard Occupational Classification (SOC) system developed by the U.S. Bureau of Labor Statistics. Then the fine grade (aka vertical) classifier utilizes job titles only for the given SOC major, limiting its classification to this vertical.

For coarse classifier, we used a proprietary implementation of the Lingo clustering algorithm. Due to high memory complexity of Lingo, we used only job titles, omitting the descriptions. Lingo applies singular value decomposition (SVD) on the TF-IDF (term frequency-inverse document frequency) term-document matrix to identify the ideal amount of clusters and cluster labels. Documents are assigned to clusters based on cosine distance.

Fine-level classifier component of Carotene is a k nearest neighbor (kNN) classifier with k empirically set to 20. We use open-source search engine library Lucene which gives us a classification response time of less than 100ms.

### 3.2 New Approaches

Word2vec (W2V), developed by Mikolov et al [9], uses shallow neural network to produce high dimensional vector representations for words and phrases. As its training objective, the neural network uses the Skip-gram model, which aims to find vector representations for each word that are useful for predicting surrounding words in a sentence or a document for that word. The relative placement of these vectors in high dimensional space turns out to be related to meanings of corresponding words. Not only vectors for similar words like "gold" and "silver" are close to each other, some semantic relationships are also preserved. For example the semantic relationship "king is to man as queen is to woman" is preserved in the vector space by satisfying this relationship $v_{king} - v_{man} \approx v_{queen} - v_{woman}$ for the corresponding vectors. However, W2V models vector representation for single word only. It doesn't learn vector representation for multiple words scenarios, e.g., sentences, documents and job titles. So given individual vector representation of every words, taking the average turns out an intuitive solution to generate one single vector representation for these words.

Besides averaging, another way to take advantage of W2V is by using Word Mower's Distance (WMD), which is a special case of the well-known Earth Mower's Distance metric and can be used to calculate the distance of documents. WMD metric is defined as the minimum possible sum of changes needed to convert one set of vectors to the other one. WMD has high time complexity: $O(p^3 \log p)$ where p is the number of unique words in documents. Kusner et al [6] present speed improvements by filtering unlikely candidates using functions that has lower time complexity and are proved to be lower-bounds to WMD. One simplistic approach for assigning a representative vector to a document is to take the average of the word vectors for the words in it, called Word Centroid Distance (WCD).

Another approach is to directly train models of paragraph vectors using artificial neural networks whose objective is to predict words in the document. Quoc Le and Tomas Mikolov [10] applied sentiment analysis and information retrieval tests to paragraph vector similarity. They found it to have significantly less classification and retrieval error compared to WCD based document similarity and other comparable algorithms.

## 4. EXPERIMENTS

### 4.1 Test Use Case

We evaluate the performance on task of job title classification by KNN, i.e., classifying query job title by top k most similar titles in our knowledge. For example, given query title "Senior Java Programmer, NY", the most similar titles found include "Entry-level Java Developer", "Matlab programmer New york", "J2EE engineer" if k = 3. Then the query title will be classified as "Java Developer" because two of the three most similar titles ("Entry-level Java Developer" and "J2EE Engineer") are normalized and labeled as "Java Developer" class.

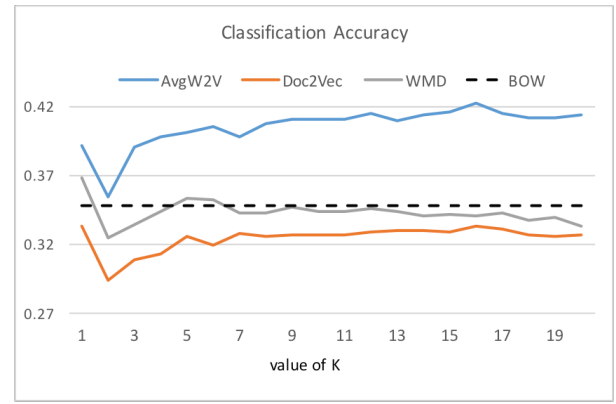

Figure 1: Accuracy of proposed methods with different k value set-up

### 4.2 Results

Our test data includes 1667 human-labeled query titles and the knowledge base contains 1002737 reference titles that fall into 5425 pre-defined occupation categories. We used accuracy metric as the evaluation metric. Figure 1 shows the performance of all methods with varying k from 1 to 20 except the current BOW (dash line) approach is evaluated only with k = 20 because it didn't parameterize k and 20 is an optimized value based on cross validation. Here are what we observed from the figure: 1) AvgW2V produced the best performance with significant gap to others. 2) W2V with EMD gave similar accuracy to BOW and 3) Doc2Vec can't beat BOW. 4) k = 1 is a good choice although k > 10 is also a safe value.

## 5. FUTURE WORK

Currently all word vectors contribute equally in building document vectors. We will try weighted version of these approaches because some words (e.g., Senior, Junior) may not be important and even introduce noise when measuring the similarity (e.g., NY, immediate hire).